\pdfoutput=1

\documentclass[11pt]{article}

\usepackage[]{EACL2023}

\usepackage{times}
\usepackage{latexsym}

\usepackage[T1]{fontenc}

\usepackage[utf8]{inputenc}
\usepackage{microtype}

\usepackage{multirow}
\usepackage{inconsolata}
\usepackage{xcolor}
\usepackage{makecell}
\usepackage{rotating}
\usepackage{amssymb}
\usepackage{pifont}
\usepackage{colortbl} 
\usepackage{algorithm}
\usepackage{algorithmic}
\usepackage{amsmath}

\usepackage{newfloat}
\usepackage{listings}
\DeclareCaptionStyle{ruled}{labelfont=normalfont,labelsep=colon,strut=off} 
\floatstyle{ruled}
\newfloat{listing}{tb}{lst}{}
\floatname{listing}{Listing}
\usepackage{xcolor}         
\usepackage{enumitem}
\newcommand{\dnn}{\textsc{DNN}}
\newcommand{\gluecons}{GLUECons}

\title{\vspace*{-0.5in}
{{\small \hfill AAAI 2023}\\
\vspace*{.25in}}GLUECons: A Generic Benchmark for Learning Under Constraints}
\author {
    Hossein Rajaby Faghihi\textsuperscript{\rm 1}, 
    Aliakbar Nafar\textsuperscript{\rm 1}, 
    Chen Zheng\textsuperscript{\rm 1},
    Roshanak Mirzaee\textsuperscript{\rm 1}, 
    Yue Zhang\textsuperscript{\rm 1},
    \\
    \textbf{Andrzej Uszok}\textsuperscript{\rm 2}, 
    \textbf{Alexander Wan}\textsuperscript{\rm 3\thanks{\hspace{1mm} Work done during internship at Michigan State University}},
    \textbf{Tanawan Premsri}\textsuperscript{\rm 1}, 
    \textbf{Dan Roth}\textsuperscript{\rm 4}, and  
    \textbf{Parisa Kordjamshidi}\textsuperscript{\rm 1} 
    \\
    \textsuperscript{\rm 1} Michigan State University,
    \textsuperscript{\rm 2} Florida Institute for Human and Machine Cognition, \\
    \textsuperscript{\rm 3} University of California Berkeley,
    \textsuperscript{\rm 4} University of Pennsylvania \\
    \{rajabyfa, nafarali, zhengc12, mirzaeem, zhan1624, premsrit, kordjams\}@msu.com, \\
    auszok@ihmc.org, 
    alexwan@berkeley.edu,
    danroth@seas.upenn.edu
}

\begin{document}
\setlength\arrayrulewidth{0.5pt}
\maketitle
\begin{abstract}
Recent research has shown that integrating domain knowledge into deep learning architectures is effective -- it helps reduce the amount of required data, improves the accuracy of the models' decisions, and improves the interpretability of models. However, the research community is missing a convened benchmark for systematically evaluating knowledge integration methods.
In this work, we create a benchmark that is a collection of nine tasks in the domains of natural language processing and computer vision. In all cases, we model external knowledge as {\it constraints}, specify the sources of the constraints for each task, and implement various models that use these constraints.
We report the results of these models using a new set of extended evaluation criteria in addition to the task performances for a more in-depth analysis. This effort provides a framework for a more comprehensive and systematic comparison of constraint integration techniques and for identifying related research challenges. It will facilitate further research for alleviating some problems of state-of-the-art neural models.

\end{abstract}

\section{Introduction}
\textbf{Deep Learning Shortcomings} Recent advancements in machine learning are proven very effective in solving real-world problems in various areas, such as vision and language. 
However, there are still remaining challenges. First, machine learning models mostly fail to perform well on complex tasks where reasoning is crucial~\cite{schubotz2018introducing} while human performance does not drop as much when more steps of reasoning are required. Second, deep neural networks~(\dnn{}s) are known to be data-hungry, making them struggle on tasks where the annotated data is scarce~\cite{li2020metamt,zoph2016transfer}. Third, models often provide results that are inconsistent~\cite{li2019logic, Gardner2020EvaluatingNM} even when they perform well on the task. Prior research has shown that even large pre-trained language models performing well on a specific task may suffer from inconsistent decisions and indicate unreliability when attacked under adversarial examples and specialized test sets that evaluate their logical consistency~\cite{Gardner2020EvaluatingNM,mirzaee2021spartqa}.
This is especially a major concern when interpretability is required~\cite{mathews2019explainable}, or there are security concerns over applications relying on the decisions of \dnn{}s~\cite{brundage2020toward}.

\noindent\textbf{Knowledge Integration Solution}
To address these challenges, one direction that the prior research has investigated is neuro-symbolic approaches as a way to exploit both symbolic reasoning
and sub-symbolic learning. Here, we focus on a subset of these approaches for the integration of external knowledge in deep learning. Knowledge can be represented through various formalisms such as logic rules~\cite{hu2016harnessing,nandwani2019primal}, Knowledge graphs~\cite{zheng2022relevant}, Context-free grammars~\cite{deutsch2019general}, Algebraic equations~\cite{stewart2017label}, or probabilistic relations~\cite{constantinou2016integrating}. A more detailed investigation of available sources of knowledge and techniques to integrate them with \dnn{}s is surveyed in \cite{von2019informed,dash2022review}.
Although integrating knowledge into \dnn{}s is done in many different forms, we focus on explicit knowledge about the latent and/or output variables. More specifically, we consider the type of knowledge that can be represented as declarative constraints imposed~(in a soft or hard way) on the models' predictions, during training or at inference time. The term knowledge integration is used in the scope of this assumption in the remainder of this paper.

\noindent\textbf{Hurdle of Knowledge Integration}
Unfortunately, most prior research on knowledge integration has only focused on evaluating their proposed method compared to baseline \dnn{} architectures that ignore the knowledge. Consequently, despite each method providing evidence of its effectiveness~\cite{hu2016harnessing, nandwani2019primal}, there is no comprehensive analysis that can provide a better understanding of the use cases, advantages, and disadvantages of methods, especially when compared with each other. 
The lack of such analysis has made it hard to apply these approaches to a more diverse set of tasks by a broader community and provide a clear comparison with existing methods. We mainly attribute this to three factors: 1) the lack of a standard benchmark with systematic baselines, 2) the difficulty of finding appropriate tasks where constraints are applicable, and 3) the lack of supporting libraries for implementing various integration techniques.

Due to these three factors, many research questions are left open for the community, such as (1) The difference in the performance of models when knowledge is integrated during inference vs. training or both, (2) The comparison of the influence of integration methods when combined with simpler vs. more complex baselines, (3) The effectiveness of training-time integration models on reducing the constraint violation, (4) The impact of data size on the effectiveness of the integration methods.

\noindent\textbf{Common Ground for Comparison} 
The contribution of this paper is providing a common ground for comparing techniques for knowledge integration by collecting a new benchmark to facilitate research in this area.
Our new benchmark, called \gluecons{}, contains a collection of tasks suitable for constraint integration, covering a spectrum of constraint complexity, from basic linear constraints such as mutual exclusivity to more complex constraints expressed in first-order logic with quantifiers. 
We organize the tasks in a repository with a unified structure where each task contains a set of input examples, their output annotations, and a set of constraints~(written in first-order logic).
We limit the scope of knowledge in \gluecons{} to logical constraints\footnote{Throughout this paper, we use the terms constraint integration, knowledge integration, or integration methods interchangeably to refer to the process of integration of knowledge into the \dnn{}s.}.

\noindent\textbf{Selected Tasks} 
\gluecons{} contains tasks ranging over five different types of problems categorized based on the type of available knowledge. 
This includes \textbf{1)} Classification with label dependencies: Mutual exclusivity in multiclass classification using  MNIST~\cite{lecun1998gradient} and Hierarchical image classification using CIFAR 100~\cite{krizhevsky2009learning}, \textbf{2)} Self-Consistency in decisions: What-If Question Answering~\cite{Tandon2019WIQAAD}, Natural Language Inference~\cite{bowman-etal-2015-large}, BeliefBank~\cite{kassner2021beliefbank},\textbf{ 3)} Consistency with external knowledge: Entity and Relation Extraction using CONLL2003~\cite{sang2003introduction},
\textbf{4)} Structural Consistency: BIO Tagging,  \textbf{5)} Constraints in (un/semi)supervised setting: MNIST Arithmetic and Sudoku.
These tasks either use existing datasets or are extensions of existing tasks, reformulated so that the usage of knowledge is applicable to them. We equip these tasks with constraint specifications and baseline results.

\noindent\textbf{Evaluation} 
For a fair evaluation and to isolate the effect of the integration technique, we provide a repository of models and code for each task in both PyTorch~\cite{NEURIPS2019_9015} and DomiKnows~\cite{faghihi2021domiknows} frameworks.
DomiKnows is an easy-to-use tool for expressing constraints in first-order logic with automatic conversion to linear constraints. It provides a modular interface for modeling and applying constraints, making it easier to consistently test different integration methods while the rest of the configurations remain unchanged.

For a more comprehensive evaluation, we introduce a set of new criteria in addition to the original task performances to measure 1) the effectiveness of the techniques in increasing the consistency with knowledge 2) the execution run-time, 3) the effectiveness of methods in the low-data regime, 4)the ability to reduce the need for complex models, and 5) the ability to express various forms of knowledge.

\noindent\textbf{Baselines}
We analyze and evaluate a set of knowledge integration methods to serve as baselines for \gluecons{}.
Our baselines cover a set of fundamentally different integration methods, where the integration is addressed either during inference or training of \dnn{}s.
\gluecons{} can be used as blueprints to highlight the importance of integrating constraints with \dnn{}s for different types of tasks and provides inspiration for building such constraints when working on new tasks.

\noindent In summary, the contributions of this paper are 1) We propose the first extensive benchmark exclusively designed for evaluating constraint integration methods in deep learning~(\gluecons{}), 2) We define new evaluation criteria in addition to the task performance for a comprehensive analysis of the techniques, and 3) We establish standard baselines for the tasks in this benchmark based on multiple constraint integration methods.

\section{Constraint Integration in Prior Research}
\label{sec:related}
Knowledge integration, often, is considered a subset of Neuro-symbolic~\cite{de2019neuro,amizadeh2020neuro,huang2021scallop} approaches that build on the intersection of neural learning and symbolic reasoning.
\citeauthor{von2019informed} surveyed prior research on knowledge integration in three directions: knowledge source, knowledge representation, and the stage of knowledge integration. \citeauthor{dash2022review} has also studied existing methods where the integration can be done through either transforming the input data, the loss function, or the model architecture itself. 
Knowledge integration has also been investigated in probabilistic learning frameworks~\cite{de2007problog,richardson2006markov,bac-jmlr17} and their modern extensions which use neural learning~\cite{deepproblog,huang2021scallop,winters2021deepstochlog}. 
Recent research has explored knowledge integration via bypassing the formal representations and expressing knowledge in the form of natural language as a part of the textual input~\cite{saeed2021rulebert,clark2020transformers}.
As of formal representations, knowledge integration has been addressed at both inference~\cite{lee2019gradient,scholak2021picard,dahlmeier2012beam} and training time~\cite{hu2016harnessing,nandwani2019primal,xu2018semantic}.

\noindent\textbf{Inference-Time Integration: }

The inference-based integration techniques optimize over the output decisions of a \dnn{}, where the solution is restricted by a set of constraints expressing the knowledge~\cite{RothYi05,chang2012structured}.

These methods aim at finding a valid set of decisions given the constraints, while their objective is specified by the output of the learning models. As a result of this fixed objective and the fact that approximation approaches are generally used to find the best solution, we expect that the type of optimization technique will not significantly affect the performance of inference-time integration methods--we will see this later in our results too. 



Prior research has investigated such integration by using variants of beam search~\cite{hargreaves2021incremental,borgeaud2020leveraging,dahlmeier2012beam}, path search algorithm~\cite{lu2021neurologic}, linear programming~\cite{RothYi05,roth2017integer,chang2012structured}, finite-state/push-down Automata~\cite{deutsch2019general}, or applying gradient-based optimization at inference~\cite{lee2019gradient,lee2017enforcing}.
We use Integer Linear Programming~(ILP)~\cite{RothYi05,roth2017integer} approach to evaluate the integration of the constraints at inference time. We choose ILP as the current off-the-shelf tool performing a very efficient search and offering a natural way to integrate constraints as far as we can find a linear form for them~\cite{faghihi2021domiknows,kordjamshidi2015saul}.

\noindent\textbf{Training-Time Integration: }
Several recent techniques have been proposed for knowledge integration at training time~\cite{nandwani2019primal,hu2016harnessing,xu2018semantic}. Using constraints during training usually requires finding a differentiable function expressing constraint violation. This will help to train the model to minimize the violations as a part of the loss function.
Integrating knowledge in the training loop of \dnn{}s is a challenging task. However, it can be more rewarding than the inference-based integration methods as it reduces the computational overhead by alleviating the need for using constraints during inference.
Although such methods cannot guarantee that the output decisions would follow the given constraints without applying further operations at inference-time, they can substantially improve the consistency with the constraints~\cite{li2019logic}.

Prior research has investigated this through various soft interpretations of logic rules~\cite{nandwani2019primal, asai2020logic}, rule-regularized supervision~\cite{hu2016harnessing,guo2021inference}, re-enforcement learning~\cite{yang2021safe}, and black-box semantic~\cite{xu2018semantic} or sampling~\cite{ahmed2022pylon} loss functions, which directly train the network parameters to output a solution that obeys the constraints.

To cover a variety of techniques based on the previous research, we select Primal-Dual~(PD)~\cite{nandwani2019primal} and Sampling-Loss~(SampL)~\cite{ahmed2022pylon} methods as baselines for our new benchmark. The PD approach relies on a soft logic interpretation of constraints, while the SampL is a black-box constraint integration. We discuss some of the existing methods in more detail in Section `Baselines.'
\subsection{Applications and Tasks}
Constraint integration has been investigated for several applications in prior research including SQL query generation~\cite{scholak2021picard}, program synthesize~\cite{austin2021program,ellis2021dreamcoder}, semantic parsing~\cite{clarke2010driving,lee2021toward}, question answering~\cite{asai2020logic}, entity and relation extraction~\cite{guo2021inference}, sentiment analysis~\cite{hu2016harnessing}, visual question answering~\cite{huang2021scallop}, image captioning~\cite{anderson2017guided}, and even text generation~\cite{lu2021neurologic}. 





\section{Criteria of Evaluation}
\label{sec:evaluation}
We extend the evaluation of the constraint integration methods beyond measuring task performance. The list of proposed evaluation criteria for such an extended comparison is as follows. 

\noindent\textbf{Individual metrics of each task:}
The first criterion to evaluate the methods is the conventional metric of each task, such as accuracy or precision/recall/F1 measures. 

\noindent\textbf{Constraint Violation:}
Even when the integration method cannot improve the model's performance, improving the consistency of its predictions will make the neural models more reliable.
A consistency measure quantifies the success of the integration method in training a neural network to follow the given constraints. We measure consistency in terms of constraint violation. We compute the ratio of violated constraints over all predicted outputs. A smaller number indicates fewer constraint violations and, consequently, a higher consistency with the available knowledge.

\noindent\textbf{Execution Run-Time:}
Another critical factor in comparing the constraint integration methods is the run-time overhead. This factor becomes even more critical when the integration happens during inference. This criterion helps in analyzing the adequacy of each technique for each application based on the available resources and the time sensitivity of the decision-making for that application. We measure this evaluation criteria by simply computing the execution time of each integration method both during training and inference. This metric can reflect the overhead of each integration method more accurately by taking into account the new parameters that should be optimized and the additional computations with respect to the complexity of the constraints.



\noindent\textbf{Low-data vs full-data performance:}
For many problems, there is no large data available either due to the high cost or infeasibility of obtaining labeled data.
Integrating constraints with deep neural learning has been most promising in such low-resource settings~\cite{nandwani2019primal,guo2021inference}. We measure the improvement resulting from the integration methods on both low and full data. This evaluation will help in choosing the most impactful integration method based on the amount of available data when trying to apply integration methods to a specific task.  

\noindent\textbf{Simple baseline vs Complex baseline:}
An expected impact of constraint integration in \dnn{}s is to alleviate the need for a large set of parameters and achieve the same performance using a smaller/simpler model. Additionally, it is important to evaluate whether the integration method can only affect the smaller network or the very large SOTA models can be improved too. This will indicate whether large networks/pre-trained models can already capture the underlying knowledge from the data or explicit constraint integration is needed to inject such knowledge. In addition to the number of parameters, this metric also explores whether knowledge integration can reduce the need for pre-training. This is especially important for the natural language domain, where large pre-trained language models prevail.

\noindent\textbf{Constraint Complexity:}
This criterion evaluates the limitations of each method for integrating different types of knowledge. Some methods consider the constraints a black box with arbitrary complexity, while others  may only model a specific form of constraint. This criterion specifies the form/complexity of the constraints that are supported by each technique. To evaluate this, we characterize a set of constraint complexity levels and evaluate whether each technique can model such constraints. 

\section{Selected Tasks}
\gluecons{} aims to provide a basis for comparing constraint integration methods. We have selected/created a collection of tasks where constraints can potentially play an important role in solving them. We provide five different problem categories containing a total of nine tasks. More details of tasks' constraints are available in the Appendix. This collection includes a spectrum of very classic tasks for structured output prediction, such as multi-class classification to more involved structures and knowledge, such as entity relation extraction and Sudoku.




\subsection{Classification with Label Dependency}


\noindent\textbf{Simple Image Classification.}
In this task, we utilize the classic MNIST~\cite{deng2012mnist} dataset and classify images of handwritten digits in the range of $0$ to $9$. 
The constraint used here is the mutual exclusivity of the ten-digit classes. Each image can only have one valid digit label as expressed in the following constraint,
\begin{equation*}
    \text{IF } digit_i(x) \Rightarrow \neg \vee_{j=!i}^{j\in[0-9]}digit_j(x),
\end{equation*}
where $digit_i(x)$ is $1$ if the model has predicted $x$ to be an image representing the digit $i$.  This task is used as a basic validation of the constraint integration methods, though it is not very challenging and can also be addressed by a ``Softmax'' function.

\noindent\textbf{Hierarchical Image Classification.}
The hierarchical relationships between labels present a more complex label dependency in multi-label and multi-class tasks. 
We use the CIFAR-100~\cite{krizhevsky2012imagenet}, which includes 100 image classes, each belonging to 20 parent classes forming a hierarchical structure. This dataset with 60k images is an extension of the classic CIFAR-10~\cite{krizhevsky2012imagenet}. To create a smaller dataset, we select 10\% of these 60k images. For this task, the output is a set of labels for each image, including one label for each level. The constraints are defined as,

\begin{equation*}
    \text{IF } L_1 \subset L_2 : L_1(x) \Rightarrow L_2(x),
\end{equation*} 
where $L_1$ and $L_2$ are labels, $L_1(x)$ is $True$ only if the models assigns label $L_1$ to $x$, and $L_1 \subset L_2$ indicates that $L_1$ is a subclass of $L_2$.

\subsection{Self Consistency in Decisions}
\dnn{}s are subject to inconsistency over multiple decisions while being adept at answering specific questions~\cite{camburu2019make}. Here, we choose three tasks to evaluate whether constraints help ensure consistency between decisions.

\noindent\textbf{Causal Reasoning.}
WIQA~\cite{Tandon2019WIQAAD} is a question-answering (QA) task that aims to find the line of causal reasoning by tracking the causal relationships between cause and effect entities in a document.
The dataset includes $3993$ questions. 
Following the work by \cite{asai2020logic}, we impose symmetry and transitivity constraints on the sets of related questions.
For example, the symmetry constraint is defined as follows:
$symmetric(q, \neg q) \Rightarrow F(q,C) \land \neg F(\neg q,C)$
where $q$ and $\neg q$ represent the question and its negated variation, $C$ denotes the document, and $\neg F$ is the opposite of the answer $F$.

\noindent\textbf{Natural Language Inference.}
Natural Language Inference~(NLI) is the task of evaluating a hypothesis given a premise, both expressed in natural language text. Each example contains a premise ($p$), hypothesis ($h$), and a label/output ($l$) which indicates whether $h$ is ``entailed,'' ``contradicted'', or ``neutral'' by $p$. 

Here, we evaluate whether NLI models benefit from consistency rules based on logical dependencies. We use the SNLI \cite{bowman-etal-2015-large} dataset, which includes 500k examples for training and 10k for evaluation. Furthermore, we include $A^{ESIM}_{1000}$~\cite{minervini2018adversarially}, which is an augmented set over the original dataset containing more related hypotheses and premise pairs to enforce the constraints. Four consistency constraints (symmetric/inverse, transitive) are defined based on the (Hypothesis, Premise) pairs. An example constraint is as follows:
\begin{equation*}
    \operatorname{neutral}\left(h,p\right)\Rightarrow\neg \operatorname{contradictory}\left(p, h\right),
\end{equation*}
where $\operatorname{neutral}\left(h, p\right)$ is \textit{True} if $h$ is undetermined given $p$. The complete constraints are described in~\cite{minervini2018adversarially}.


\noindent\textbf{Belief Network Consistency.} The main goal of this task is to impose global belief constraints to persuade models to have consistent beliefs. As humans, when we reason, we often rely upon our previous beliefs about the world, whether true or false. We can always change our minds about previous information based on new information, but new beliefs should not contradict previous ones. Here, entities and their properties are used as facts. We form a global belief network that must be consistent with those derived from a given knowledge base. We use Belief Bank~\cite{kassner2021beliefbank} dataset to evaluate the consistency perseverance of various techniques. The dataset consists of $91$ entities and $23k$~($2k$ train, $1k$ dev, $20k$ test) related facts extracted from ConceptNet~\cite{DBLP:journals/corr/SpeerCH16}. There are $4k$ positive and negative implications between the facts in the form of a constraint graph. For example, the fact ``Is a bird'' would imply ``can fly,'' and the fact ``can fly'' refute the fact ``Is a dog''. Formally, the constraints are defined as follows:
\vspace{-2mm}
\begin{align*}
    \forall F_1,F_2 \in \text{Facts};& \\
    \text{IF}\ F_1,F_2 &\in \text{Pos Imp} \Rightarrow \neg F_1(x) \lor F_2(x)
    \\ \text{IF}\ F_1,F_2 &\in \text{Neg Imp} \Rightarrow \neg F_1(x) \lor \neg F_2(x),
\end{align*}
``Pos Imp'' means a positive implication.

\subsection{Consistency with External Knowledge}
This set of tasks evaluates the constraint integration methods in applying external knowledge to the \dnn{}s' outputs. 

\noindent\textbf{Entity Mention and Relation Extraction~(EMR).}
This task is to extract entities and their relationships from a document. Here, we focus on the CoNLL2003~\cite{sang2003introduction} dataset, which contains about $1400$ articles.  
There are two types of constraints involved in this task: 1) mutual exclusivity between entity/relationship labels and 2) a restriction on the types of entities that may engage in certain relationships.
An example constraint between entities and relationship types is as follows: 
\begin{equation*}
    \operatorname{IF } \operatorname{Work\_for}(x1,x2) \Rightarrow
    \operatorname{Person}(x1) \land \operatorname{Org}(x2),
\end{equation*}
where $\operatorname{Predicate}(x)$ is $True$ if the network predicted input $x$ to be of type $\operatorname{Predicate}$.

\subsection{Structural Consistency}
In this set of tasks, we evaluate the impact of constraint integration methods in incorporating structural knowledge over the task's outputs.

\noindent\textbf{BIO Tagging.}
The BIO tagging task aims to identify spans in sentences by tagging each token with one of the ``Begin,'' ``Inside,'' and ``Outside'' labels.
Each tagging output belongs to a discrete set of BIO tags $T \in $ [`O', `I-*', `B-*'], where `*' can be any type of entity. Words tagged with O are outside of named entities, while the `B-*' and `I-*' tags are used as an entity's beginning and inside parts. 
We use the CoNLL-2003~\cite{sang2003introduction} benchmark to evaluate this task. This dataset includes $1393$ articles and $22137$ sentences. The constraints of the BIO tagging task are valid BIO sequential transitions; for example, the ``before'' constraint is defined as follows: 
\begin{equation*}
    \operatorname{If } I(x_i+1) \xrightarrow{} B(x_i),
\end{equation*}
where `B-*' tag should appear before `I-*' tag. $x_i$ and $x_{i+1}$ are any two consecutive tokens.

\subsection{Constraints in (Un/Semi) Supervised Learning}
We select a set of tasks for which the constraints can alleviate the need for direct supervision and provide a distant signal for training \dnn{}s. 

\noindent\textbf{Arithmetic Operation as Supervision for Digit Classification.} 
We use the MNIST Arithmetic~\cite{bloice2020performing} dataset. The goal is to train the digit classifiers by receiving supervision, merely, from the sum of digit pairs. For example, for image pairs of $5$ and $3$ in the training data, we only know their labels' sum is $8$.
This dataset is relatively large, containing 10k image pairs for training and 5k for testing.
This task's constraint forces the networks to produce predictions for pairs of images where the summation matches the ground-truth sum.
The following logical expression is an example constraint for this task:
\begin{align*}
    &S(\{img_1, img_2\}) \Rightarrow \\
    &\bigvee_{M=max(0,S-9)}^{M=min(S,9)} M(img_1) \wedge \{S-M\}(img_2),
\end{align*}
where $S(\{img_1, img_2\})$ indicates that the given summation label is $S$ and $M(img_i)$ indicates that the $i$th image has the label $M$.

\noindent\textbf{Sudoku.}
This task evaluates whether constraint integration methods can help \dnn{}s to solve a combinatorial search problem such as Sudoku. Here, integration methods are used as an inference algorithm with the objective of solving one Sudoku, while the only source of supervision is the Sudoku constraints. As learning cannot be generalized in this setting, it should be repeated for each input. The input is one Sudoku table partially filled with numbers, and the task is to fill in a number in each cell such that: 
"There should not be two cells in each row/block/column, with the same value" or formally defined as:
\begin{align*}
    &\text{IF } \operatorname{digit_i}(x) \land \\&(\operatorname{same\_row}(x, y) \lor \operatorname{same\_col}(x,y) \lor \operatorname{same\_block}(x,y)) \\
    & \Rightarrow \neg \operatorname{digit_i}(y),
\end{align*}
where $x$ and $y$ are variables regarding the cells of the table, $i \in [0, n]$ for a $n*n$ Sudoku, $digit_i(x)$ is $True$ only if the value of $x$ is predicted to be $i$.
For this task, we use an incomplete $9*9$ Sudoku for the full-data setting and a $6*6$ Sudoku representing the low-data setting. 

\section{Baselines}
\label{sec:baselines}
For constraints during training, we use two approaches.

\noindent\textbf{Primal-Dual~(PD).} This approach~\cite{nandwani2019primal} converts the constrained optimization problem into a min-max optimization with Lagrangian multipliers for each constraint and augments the original loss of the neural models. This new loss value quantifies the amount of violation according to each constraint by means of a soft logic surrogate. During training, they optimize the decision by minimizing the original violation, given the labels, and maximizing the Lagrangian multipliers to enforce the constraints. It is worth noting that all related work in which constraint violation is incorporated as a regularization term in the loss objective follows very similar variations of a similar optimization formulation.

\noindent\textbf{Sampling-Loss (SampL).} This approach~\cite{ahmed2022pylon} is an extension of the semantic loss~\cite{xu2018semantic} where instead of searching over all the possibilities in the output space to find satisfying cases, it randomly generates a set of assignments for each variable using the probability distribution  of the neural network's output. The loss function is formed as:
\begin{equation*}
    L^{S}(\alpha, p)=\frac{\sum_{x^{i} \in X \land x^{i} \models \alpha} p\left(x^{i} \mid p\right)}{\sum_{x^{i} \in X} p\left(x^{i} \mid p\right)},
\end{equation*}
where $X$ is the set of all possible assignments to all output variables, and $x^{i}$ is one assignment. Here, $\alpha$ is one of the constraints.

To utilize the constraints during prediction, we use the following approaches.

\noindent\textbf{Integer Linear Programming.}~(ILP)~\cite{RothYi05} is used to formulate an optimization objective in which we want to find the most probable solution for $\log{F(\theta)}^\top y$, subject to the constraints. Here, $y$ is the unknown variable in the optimization objective, and $F(\theta)$ is the network's output probabilities for each variable in $y$. The constraints on $y$ are formulated as $\mathcal{C}\left(y\right)\le0$.

\noindent\textbf{Search and Dynamic Programming.} for some of the proposed benchmarks, when applicable, we use the $A^*$ search or Viterbi algorithm to choose the best output at prediction time given the generated probability distribution of the final trained network~\cite{lu2021neurologic}.


\begin{table}[t]
\scriptsize
\begin{tabular}{l|l|l}
\hline
     \textbf{Task}& \textbf{Strong Baseline} & 
     \begin{tabular}[l]{@{}l@{}}
     \textbf{Simple Baseline} \end{tabular}  \\ \hline
     
     Image Cls. & CNN $+$ MLP
     & MLP \\\hline
     Hier. Image Cls.  & Resnet18 $+$ MLP & - \\ \hline
     NLI  &  RoBERTa $+$ MLP &  - \\ \hline
     
     Causal Rea. & RoBERTa $+$ MLP & BERT $+$ MLP \\ \hline
     
     BIO Tagging & BERT $+$ MLP & Bi-LSTM $+$ MLP \\ \hline
     
     NER & W$2$V $+$ BERT $+$ MLP & W$2$V $+$ Bi-LSTM $+$ MLP\\ \hline
     
     Ari. Operation & CNN $+$ MLP &  -\\ \hline
     
     BeliefNet Con.  & RoBERTa + MLP & Word vectors + MLP \\ \hline
     
     Sudoku &
     \begin{tabular}[l]{@{}l@{}} ($n*n*n$) Vector \\ directly learns probabilities \end{tabular} 
     & - \\
\hline
\end{tabular}
\vspace{-1mm}
\caption{\small Baselines for each task. The basic models we used are RoBERTa~\cite{Liu2019RoBERTaAR}, BERT~\cite{devlin2018bert}, W2V~\cite{mikolov2013efficient}, CNN \cite{lecun1998gradient}, and MLP. The simple baseline means fewer parameters. [\textbf{KEYS}: \textbf{Cls.}$=$classification, \textbf{Hier.}$=$Hierarchical,
\textbf{NLI}$=$ Natural Language Inference,
\textbf{Rea.}$=$reasoning, \textbf{Ari.}$=$Arithmetic, \textbf{BeliefNet}$=$Belief Network, \textbf{Con.}$=$Consistency]. For the Sudoku, the model is not a generalizable DNN and the method uses the integration methods as an inference algorithm to solve one specific table.\vspace{-2mm}}
\label{tab:baselines}
\end{table}

\section{Experiments and Discussion}
\label{sec:experiments}
This section highlights our experimental findings using proposed baselines, tasks, and evaluation criteria. Details on experimental designs, training hyper-parameters, codes, models, and results can be found on our website\footnote{\url{https://hlr.github.io/gluecons/}}. The basic architectures for each task are shown in Table \ref{tab:baselines}.
The results of the experiments are summarized in Table \ref{tab:results}. The columns represent evaluation criteria, and the rows represent tasks and their baselines. Each task's model `row' records the strong/simple baseline's performance without constraint integration, and below that, the improvements or drops in performance after adding constraints are reported. Here, we summarize the findings of these experiments by answering the following questions. 

\begin{table}[t]
\centering
\scriptsize
\begin{tabular}{lcccccc}
\hline
&
\tabular{@{}c@{}}Mutual \\Excl.\endtabular
 &
\tabular{@{}c@{}}Seq. \\structure\endtabular  &
\tabular{@{}c@{}}Lin. \\Const\endtabular &
\tabular{@{}c@{}}Log. \\Const\endtabular &
\tabular{@{}c@{}}Log Const \\ + quantifier\endtabular &
\tabular{@{}c@{}}Prog\\ Const\endtabular
 \\
 \hline

Softmax & \textcolor{olive}{\LARGE $\checkmark$}&   \textcolor{red}{\Large \ding{55}}&   \textcolor{red}{\Large \ding{55}}&  \textcolor{red}{\Large \ding{55}} &  \textcolor{red}{\Large \ding{55}} & \textcolor{red}{\Large \ding{55}}\\
PD      &  \textcolor{olive}{\LARGE $\checkmark$} &  \textcolor{olive}{\LARGE $\checkmark$} &  \textcolor{olive}{\LARGE $\checkmark$} &  NC &  NC & \textcolor{red}{\Large \ding{55}}\\
SampL   &  \textcolor{olive}{\LARGE $\checkmark$} &  \textcolor{olive}{\LARGE $\checkmark$} &  \textcolor{olive}{\LARGE $\checkmark$} &  \textcolor{olive}{\LARGE $\checkmark$} &  \textcolor{olive}{\LARGE $\checkmark$} & \textcolor{olive}{\LARGE $\checkmark$}\\
ILP     &  \textcolor{olive}{\LARGE $\checkmark$} &  \textcolor{olive}{\LARGE $\checkmark$} &  \textcolor{olive}{\LARGE $\checkmark$} &  NC &  NC & \textcolor{red}{\Large \ding{55}}\\
$A^*$ \tiny{Search}    &  \textcolor{olive}{\LARGE $\checkmark$} &  \textcolor{olive}{\LARGE $\checkmark$} &  NG & NG &  NG & \textcolor{red}{\Large \ding{55}}\\
\hline
\end{tabular}
\vspace{-2mm}
\caption{The limitation of integration methods based on different types of constraints. [\textbf{KEYS}: \textbf{NC}$=$Needs Conversion, \textbf{NG}$=$No Generalization,
\textbf{Excl.}$=$ Exclusivity,
\textbf{Seq.}$=$Sequential, \textbf{Lin.}$=$Linear, \textbf{Log.}$=$Logical, \textbf{Const.}=Constraint, \textbf{Prog Const}$=$Any Constraints encoded as a program.]}
\label{tab:expressive}
\vspace{-5mm}
\end{table}

\noindent\textbf{What are the key differences in the performance of inference-time and training-time integration?}
Notably, using ILP only in inference time outperformed other baselines in most of the tasks. However, it fails to perform better than the training-time integration methods when the base model is wildly inaccurate in generating the probabilities for the final decisions. This phenomenon happened in our experiments in the semi-supervised setting and can be seen when comparing rows [\#44, \#45] to \#46. In this case, inference alone cannot help correct the model, and global constraints should be used as a source of supervision to assist with the learning process.


ILP performs better than the training-time methods when applied to simpler baselines~(see column simple baseline performance). 
However, the amount of improvement does not differ significantly when applying ILP to the simpler baselines compared to the strong ones. Additionally, the training-time methods perform relatively better on simpler baselines than the strong ones~(either the drop is less or the improvement is higher)~(compare columns `Strong Baseline` and `Simple Baseline` for `+PD` and `+SampL` rows.)

\begin{table*}[!hbt]
\small
\resizebox{1.0\textwidth}{!}{
\begin{tabular}{|l|c|c|l|c|c|cc|c|cc|}
\hline
 & & & & & & & & & &   \\
 \multirow{2}{*}{}& \multirow{2}{*}{Tasks}&\multirow{2}{*}{$\#$} & \multirow{2}{*}{Models}&
 \multirow{2}{*}{\shortstack{Strong Baseline \\ Performance}} & \multirow{2}{*}{\tabular{@{}c@{}}Simple Baseline \\Performance\endtabular}
 &\multicolumn{2}{c|}{Low data} &
 \multirow{2}{*}{\tabular{@{}c@{}}Constraint \\Violation*\endtabular} &
 \multicolumn{2}{c|}{Run-Time$^{ms}$}  \\
 & & & &  & &Size &Performance & &Training &Inference  \\\hline
\multirow{10}{*}{
\begin{turn}{90}
{\shortstack{Classification with \\ Label Dependency}} 
\end{turn}
}  
 &  \multirow{6}{*}{\tabular{@{}c@{}}Simple \\Img Cls$^{F1}$ \endtabular} &
  1 &Model & $94.23\%$ &$87.34\%$& \multirow{6}{*}{$5\%$} & $88.78\%$ & $7.17\%$ & $34$ & $27.5$ \\
 & & 2& $+$ PD & \textcolor{teal}{$\uparrow$}$0.14\%$  & \textcolor{red}{$\downarrow$}$1.14\%$  & & \textcolor{teal}{$\uparrow$}$4.40\%$ & $8.32\%$ & $36.6$ &-  \\
 & & 3&$+$ SampL & \textcolor{red}{$\downarrow$}$1.17\%$& \textcolor{teal}{$\uparrow$}$0.49\%$& & \textcolor{teal}{$\uparrow$}$3.19\%$ & $9.04\%$& $39$ &-  \\
 &  & 4&$+$ ILP & \textcolor{teal}{$\uparrow$}$0.24\%$ & \textcolor{teal}{$\uparrow$}$1.60\%$ & & \textcolor{teal}{$\uparrow$}$1.70\%$ &- &- & $31.5$ \\
&  & 5&$+$ SampL $+$ ILP & \textcolor{red}{$\downarrow$}$0.52\%$ & \textcolor{teal}{$\uparrow$}$2.02\%$ & & \textcolor{teal}{$\uparrow$}$4.40\%$ &- &- & - \\
 &  & 6&$+$ PD $+$ ILP & \textcolor{teal}{$\uparrow$}$0.32\%$ & \textcolor{teal}{$\uparrow$}$0.39\%$ & & \textcolor{teal}{$\uparrow$}$4.40\%$ &- &- & - \\
\cline{2-11}
 &\multirow{4}{*}{\tabular{@{}c@{}}Hierarchical \\Img Cls$^{F1}$ \endtabular} & 7 &Model & $58.03\%$& $52.54\%$ &\multirow{4}{*}{$10\%$} & $31.33\%$ & $39.26\%$ & $55.3$ & $48.43$ \\
  & & 8&$+$ SampL & \textcolor{teal}{$\uparrow$}$0.39\%$ & \textcolor{teal}{$\uparrow$}$0.54\%$ & &\textcolor{teal}{$\uparrow$}$2.18\%$ & $36.57\%$ &- & $55.2$ \\
   & & 9&$+$ ILP & \textcolor{teal}{$\uparrow$}$2.88\%$ & \textcolor{teal}{$\uparrow$}$3.18\%$& &\textcolor{teal}{$\uparrow$}$1.90\%$ &- &- & $55.2$ \\
   & & 10&$+$ SampL $+$ ILP & \textcolor{teal}{$\uparrow$}$2.42\%$ & \textcolor{teal}{$\uparrow$}$3.52\%$& &\textcolor{teal}{$\uparrow$}$3.82\%$ &- &- & $55.2$ \\
\hline

\multirow{18}{*}{
\begin{turn}{90}
{\tabular{@{}c@{}}Self Consistency in \\Decision Dependency\endtabular} 
\end{turn}
} &\multirow{6}{*}{\tabular{@{}c@{}}Causal$^{A}$ \\Reasoning\endtabular}
&12 & Model & $74.77\%$ & $73.80\%$ & \multirow{6}{*}{$30\%$} & $60.49\%$ & $8.60\%$  & $104$ & $46.2$ \\
 &  & 13 &$+$ PD & \textcolor{teal}{$\uparrow$}$2.17\%$  & \textcolor{teal}{$\uparrow$}$1.98\%$ & & \textcolor{teal}{$\uparrow$}$1.10\%$ & $11.36\%$ & $118.1$ & -  \\
 &  & 14 &$+$ SampL & \textcolor{teal}{$\uparrow$}$2.54\%$  & \textcolor{teal}{$\uparrow$}$2.17\%$ & & \textcolor{teal}{$\uparrow$}$1.63\%$ & $4.37\%$ & $119.4$ & -  \\
 & & 15&$+$ ILP & \textcolor{teal}{$\uparrow$}$4.03\%$ & \textcolor{teal}{$\uparrow$}$4.51\%$ & & \textcolor{teal}{$\uparrow$}$1.88\%$ & - & - & $59.2$ \\
  & & 16&$+$ SampL $+$ ILP & \textcolor{teal}{$\uparrow$}$4.15\%$ & \textcolor{teal}{$\uparrow$}$4.25\%$ & & \textcolor{teal}{$\uparrow$}$2.11\%$ & - & - & - \\
   & & 17&$+$ PD $+$ ILP & \textcolor{teal}{$\uparrow$}$3.60\%$ & \textcolor{teal}{$\uparrow$}$4.30\%$ & & \textcolor{teal}{$\uparrow$}$1.76\%$ & - & - & - \\
\cline{2-11}
 &\multirow{6}{*}{NLI$^{A}$} 
  & 18 &Model & $74.00\%$ & -&\multirow{6}{*}{$10\%$} & $68.65\%$ & $9.48\%$& $29.2$ & $10.7$  \\
 & & 19 &$+$ PD & \textcolor{teal}{$\uparrow$}$0.25\%$ &- & & \textcolor{teal}{$\uparrow$}$3.25\%$ & $7.26\%$& $31.7$ &-  \\
 & & 20&$+$ SampL & \textcolor{teal}{$\uparrow$}$0.55\%$ &- & & \textcolor{teal}{$\uparrow$}$0.95\%$ & $5.00\%$& $29.8$ & -  \\
 & & 21 &$+$ ILP & \textcolor{teal}{$\uparrow$}$8.90\%$ & - & & \textcolor{teal}{$\uparrow$}$7.75\%$ & - & - & $14.3$ \\
  & & 22 &$+$ SampL $+$ ILP & \textcolor{teal}{$\uparrow$}$8.20\%$ & - & & \textcolor{teal}{$\uparrow$}$7.05\%$ & - & - & - \\
   & & 23 &$+$ PD $+$ ILP & \textcolor{teal}{$\uparrow$}$8.75\%$ & - & & \textcolor{teal}{$\uparrow$}$10.1\%$ & - & - & - \\
\cline{2-11}

 &\multirow{6}{*}{\tabular{@{}c@{}}Belief$^{F1}$ \\Network\endtabular} & 24 &Model & $94.90\%$ & $84.46\%$& \multirow{6}{*}{$25\%$}& $94.36\%$ & $0.22\%$ & $8.3$& $7.57$ \\
 & & 25 &$+$ PD & \textcolor{teal}{$\uparrow$}$0.94\%$ & \textcolor{teal}{$\uparrow$}$0.87\%$&  & \textcolor{red}{$\downarrow$}$0.49\%$ & $0.16\%$ & $23.59$ & - \\
 & & 26 & $+$ SampL &\textcolor{red}{$\downarrow$}$0.29\%$ & \textcolor{red}{$\downarrow$}$0.95\%$ & & \textcolor{red}{$\downarrow$}$3.03\%$ & $0.01\%$& $8.5$ & - \\
 & & 27 &$+$ ILP & \textcolor{teal}{$\uparrow$}$0.21\%$ & \textcolor{red}{$\downarrow$}$0.10\%$& & \textcolor{red}{$\downarrow$}$0.97\%$ & - & - & $11$ \\
 & & 28 &$+$ SampL $+$ ILP & \textcolor{teal}{$\uparrow$}$1.10\%$ & \textcolor{red}{$\downarrow$}$3.19\%$& & \textcolor{red}{$\downarrow$}$2.31\%$ & - & - & - \\
& & 29 &$+$ PD $+$ ILP & \textcolor{teal}{$\uparrow$}$2.68\%$ & \textcolor{teal}{$\uparrow$}$1.60\%$& & \textcolor{teal}{$\uparrow$}$0.51\%$ & - & - & - \\
\hline

\multirow{6}{*}{
\begin{turn}{90}
{\tabular{@{}c@{}}Consistency \\with EK \endtabular} 
\end{turn}}
 &\multirow{6}{*}{\tabular{@{}c@{}}EMR$^{F1}$\endtabular} 
 & 30 &Model & $90.15\%$ & $85.22\%$ &\multirow{6}{*}{$20\%$} & $82.00\%$ & $1.17\%$ &$210$ &$200$  \\
 & & 31 &$+$ PD & \textcolor{red}{$\downarrow$}$1.00\%$ & \textcolor{red}{$\downarrow$}$0.30\%$ & & \textcolor{teal}{$\uparrow$}$2.42\%$ & $0.94\%$ & $245$ &-  \\
 & & 32 &$+$ SampL & \textcolor{red}{$\downarrow$}$0.30\%$ & \textcolor{teal}{$\uparrow$}$0.50\%$ & & \textcolor{teal}{$\uparrow$}$3.36\%$  &$0.98\%$ &$280$ &-  \\
 & & 33 &$+$ ILP & \textcolor{teal}{$\uparrow$}$3.02\%$ & \textcolor{teal}{$\uparrow$}$4.10\%$ & &\textcolor{teal}{$\uparrow$}$8.86\%$ & -&- &$226$  \\
 & & 34 &$+$ SampL $+$ ILP & \textcolor{teal}{$\uparrow$}$2.40\%$ & - & &\textcolor{teal}{$\uparrow$}$7.83\%$ & -&- &-  \\
 & & 35 &$+$ PD $+$ ILP & \textcolor{teal}{$\uparrow$}$1.64\%$ & - & &\textcolor{teal}{$\uparrow$}$8.15\%$ & -&- &-  \\
\hline

\multirow{7}{*}{
\begin{turn}{90}
{\tabular{@{}c@{}}Structural \\Consistency\endtabular} 
\end{turn}
} 

&\multirow{7}{*}{\tabular{@{}c@{}}BIO$^{F1}$ \\Tagging\endtabular}
 & 36 & Model & $89.56\%$ & $82.77\%$ &\multirow{7}{*}{$30\%$} & $75.36\%$ & $2.19\%$ & $361.2$ & $263.2$  \\
 & & 37 & $+$ PD & \textcolor{teal}{$\uparrow$}$0.97\%$ & \textcolor{teal}{$\uparrow$}$0.04\%$ & & \textcolor{teal}{$\uparrow$}$1.25\%$ & $0.99\%$ & $389.1$  & - \\
 & & 38 & $+$ SampL & \textcolor{red}{$\downarrow$}$0.17\%$ & \textcolor{teal}{$\uparrow$}$0.73\%$ & & \textcolor{teal}{$\uparrow$}$2.62\%$ & $0.16\%$ & $429.8$  & -  \\
& & 39 & $+$ ILP & \textcolor{teal}{$\uparrow$}$0.61\%$ & \textcolor{teal}{$\uparrow$}$2.96\%$ & & \textcolor{teal}{$\uparrow$}$3.01\%$ & - & -  & $312$  \\
& & 40 &$+$ SampL $+$ ILP & \textcolor{teal}{$\uparrow$}$0.08\%$ & \textcolor{teal}{$\uparrow$}$2.43\%$ & & \textcolor{teal}{$\uparrow$}$2.80\%$ & - & -  & - \\
& & 41 &$+$ PD $+$ ILP & \textcolor{teal}{$\uparrow$}$1.07\%$ & \textcolor{teal}{$\uparrow$}$1.83\%$ & & \textcolor{teal}{$\uparrow$}$2.73\%$ & - & -  & -  \\
& & 42 & $+ A^*$ search & \textcolor{teal}{$\uparrow$}$0.59\%$ & \textcolor{teal}{$\uparrow$}$2.97\%$ & & \textcolor{teal}{$\uparrow$}$3.03\%$ & - & - & -  \\
 \hline
\multirow{8}{*}{
\begin{turn}{90}
{\tabular{@{}c@{}}Constraints in \\(Un/Semi)Supervision\endtabular}
\end{turn}
} 
&\multirow{5}{*}{\shortstack{Arithmetic \\ Supervision for \\Digit \\ Classification$^{A}$}} 
& 43 &Model & $9.01\%$ & - &\multirow{5}{*}{$5\%$} & $10.32\%$& $96.92\%$ & $13.6$& - \\
 & & 44 &$+$ PD & \textcolor{teal}{$\uparrow$}$89.39\%$ & -  & & \textcolor{teal}{$\uparrow$}$85.01\%$& $3.18\%$&$197$ & - \\
 & & 45 &$+$ SampL & \textcolor{teal}{$\uparrow$}$89.55\%$&  - & & \textcolor{teal}{$\uparrow$}$85.60\%$ & $2.86\%$& $90.2$&  -\\
 & & 46 &$+$ ILP & \textcolor{red}{$\downarrow$}$2.11\%$ & - &   & $0.00\%$& - & - & - \\
 & & 47&$+$Supervised & \textcolor{teal}{$\uparrow$}$89.53\%$& -  & & \textcolor{teal}{$\uparrow$}$84.30\%$& $2.86\%$ & $12.5$& - \\
\cline{2-11}
 &\multirow{3}{*}{Sudoku$^{CS}$}& 48 &PD & $96.00\%$&- & \multirow{3}{*}{
 \tabular{@{}c@{}} $6*6$\\Table\endtabular} & $100\%$& $3.7\%$ &-&- \\
  & & 49 &SampL & $87.00\%$ &- &  & $100\%$ &$18.88\%$&-& - \\
 & & 50 &ILP & $100\%$& -& & $100\%$ &- &- & - \\
\hline
\end{tabular}
}
\vspace{-2mm}
\caption{Impact of constraint integration. F1, A, and CS are F1-measure, accuracy, and constraint satisfaction metrics to evaluate models' performance. The full data of the Sudoku task is a $9*9$ table. *: Constraint Violation is on the full data with the strong baselines. ms: Run-Time is computed per example/batch and is reported in milliseconds. \textcolor{teal}{$\uparrow$} indicates improvement over the initial Model performance. \textcolor{red}{$\downarrow$} indicates a drop in the performance. Run times are recorded on a machine with Intel
Core i9-9820X (10 cores, 3.30 GHz) CPU and Titan RTX
with NVLink as GPU.[\textbf{KEYS}: EK=external knowledge] }
\label{tab:results}
\end{table*}

\noindent\textbf{How does the size of data affect the performance of the integration techniques?} 
The integration methods are exceptionally effective in the low-data regime when the constraints come from external knowledge or structural information. This becomes evident when we compare the results of `EMR' and `BIO tagging' with the `Self Consistency in Decision Dependency' tasks in column `Low data/ Performance'. This is because such constraints can inject additional information into the models, compensating for the lack of training data. 
However, when constraints are built over the self-consistency of decisions, they are less helpful in low-data regimes~(rows \#12 to \#29), though a positive impact is still visible in many cases. This observation can be justified since there are fewer applicable global constraints in-between examples in the low-data regime. Typically, batches of the full data may contain tens of relationships leading to consistency constraints over their output, while batches of the low data may contain fewer relationships. The same observation is also seen as batch sizes for training are smaller.



\noindent\textbf{Does constraint integration reduce the constraint violation?}
Since our inference-time integration methods are searching for a solution consistent with the constraints, they always have a constraint violation rate of 0\%. However, training-time integration methods cannot fully guarantee the consistency. However, it is worth noting these methods have successfully reduced the constraint violation in our experiments even when the performance of the models is not substantially improved or is even slightly hurt~(see rows \#18 and \#20, rows \#24 and \#26, and rows \#30 to \#32). In general, SampL had a more significant impact than PD on making models consistent with the available task knowledge~(compare rows with `+PD' and `+SampL' in column `Constraint Violation'). 

\noindent\textbf{How do the integration methods perform on simpler baselines?}
According to our experiments, there is a significant difference between the performance of the integration methods applied to simple and strong baselines when the source of constraint was external (BIO tagging, EMR, Simple Image Cls, and Hierarchical Image Cls tasks). Moreover, we find that ILP applied to a simple baseline can sometimes achieve a better outcome than a strong model without constraints. This is, in particular, seen in the two cases of~EMR and Causal Reasoning, where the difference between the simple and strong baselines is in using a pre-trained model. Thus, explicitly integrating knowledge can reduce the need for pre-training. In such settings, constraint integration compensates for pre-training a network with vast amounts of data for injecting domain knowledge for specific tasks. 
Additionally, the substantial influence of integration methods on simple baselines compared to strong ones in these specific tasks indicates that constraint integration is more effective when knowledge is not presumably learned~(at some level) by available patterns in historical data used in the pre-raining of large language models.

\noindent\textbf{How much time overhead is added through the integration process?}
While the inference-time method~(ILP) has a computational overhead during inference, we have shown that this overhead can be minimized if a proper tool is used to solve the optimization problem~(here, we use Gurobi\footnote{\url{https://www.gurobi.com/}}.
It should be noted that training-time integration methods do not introduce additional overhead during inference; however, they typically have a high computational cost during training. 
In the case of our baselines, SampL has shown to be relatively more expensive than PD. This is because SampL has an additional cost for forming samples and evaluating the satisfaction of each sample.

\noindent\textbf{What is the effect of combining inference-time and training-time integration methods?}
Our results show that combining inference-time and training-time methods mainly yields the highest performance on multiple tasks. For example, the performance on the NLI task on low-data can yield over 10\% improvement with the combination of PD and ILP, while ILP on its own can only improve around 7\%. The rationale behind these observations needs to be further investigated. However, this can be attributed to better local predictions of the training-time integration methods that make the inference-time prediction more accurate. A more considerable improvement is achieved over the initial models when these predictions are paired with global constraints during ILP~(see rows \#16, \#28, \#29, and \#41).

\noindent\textbf{What type of constraints can be integrated using each method?}
Table~\ref{tab:expressive} summarizes the limitations of each constraint integration method to encode a specific type of knowledge. We have included ``Softmax'' in this table since it can be used to support mutual exclusivity directly in \dnn{}. However, ``Softmax'' or similar functions are not extendable to more general forms of constraints. 
SampL is the most powerful method that is capable of encoding any arbitrary program as a constraint. This is because it only needs to evaluate each constraint based on its satisfaction or violation.
A linear constraint can be directly imposed by PD and ILP methods. However, first-order logic constraints must be converted to linear constraints before they can be directly applied. Still, PD and ILP methods fail to generalize to any arbitrary programs as constraints. The $A^*$ search can generally be used for mutual exclusivity and sequential constraints, but it cannot provide a generic solution for complex constraints as it requires finding a corresponding heuristic. ~\cite{chang2012structured} show $A^*$ with constraints can be applied under certain conditions and when the feature function is decomposable.

\section{Conclusion and Future Work}
This paper presented a new benchmark, \gluecons{} for constraint integration with deep neural networks. \gluecons{} contains nine different tasks supporting a range of applications in natural language and computer vision domains. Given this benchmark, we evaluated the influence of the constraint integration methods beyond the tasks' performance by introducing new evaluation criteria that can cover the broader aspects of the effectiveness and efficiency of the knowledge integration. We investigated and compared methods for integration during training and inference. Our results indicate that, except in a few
cases, inference-time integration outperforms the training-time integration techniques, showing that training-time integration methods have yet to be fully explored to achieve their full potential in improving the DNNs. 
Our experiments show different behaviors of evaluated methods across tasks, which is one of the main contributions of our proposed benchmark.  
This benchmark can serve the research community around this area to evaluate their new techniques against a set of tasks and configurations to analyze multiple aspects of new techniques. 
In the future, we plan to extend the tasks of this benchmark to include more applications, such as spatial reasoning over natural language~\cite{mirzaee-etal-2021-spartqa}, visual question answering~\cite{huang2021scallop}, procedural reasoning~\cite{faghihi2021time,dalvi2019everything}, and event-event relationship extraction~\cite{DBLP:journals/corr/abs-2010-06727}. 

\section*{Acknowledgments}
This project is supported by National Science Foundation (NSF) CAREER award 2028626 and partially supported by the Office of Naval Research
(ONR) grant N00014-20-1-2005. Any opinions,
findings, and conclusions or recommendations expressed in this material are those of the authors and do not necessarily reflect the views of the National
Science Foundation nor the Office of Naval Research.

%
%

\bibliography{References}
\bibliographystyle{acl_natbib}

\appendix
\section{Appendix}
Here, we describe experimental details with the selected tasks and baselines.
All the execution times are recorded on a machine with Intel Core i9-9820X (10 cores, 3.30 GHz) CPU and Titan RTX with NVLink as GPU.












\subsection{Simple Image Classification}

\textbf{Task} MNIST Binary is a simple and standard dataset in which handwritten numbers from 0 to 9 appear in the form of $28*28$ pixels of an image to be classified. However, a typical model used to solve it does not predict each label independently. Designing our model in a way that predicts each class individually and independently, in combination with a low data regime, will make this dataset somewhat challenging. 

\textbf{Experiments} To solve MNIST classification, which is the benchmark for our simple image classification, we define one model per class. Each class has two CNN layers with kernel sizes of $5$, followed by a linear layer. The images are normalized with the typical values of  $0.1307$ and $0.3081$. The model trains with a batch size of $30$ for three epochs with the Adam optimizer and the learning rate of $2e-3$. At the end of the final evaluation, we show the averaged F1 measure for all the classes. The results are shown in table \ref{table:mnist_binary}.

\textbf{Simple Baseline} To create a simple baseline for this task, we flatten the images to form a vector of size $1*784$ instead of a $28*28$ image. Then we input this vector to a single-layer MLP. In this simple baseline, spatial data will be lost, but due to the simplicity of the dataset, the results are still near $90\%$.

\begin{table*}[!ht] 

\begin{center}
\begin{tabular}{ c c c c c c c c }
\hline
   Data Scale & Baseline & Baseline+ILP & SampleLoss & SampleLoss+ILP & primal-dual & primal-dual+ILP \\ 
 $100\%$   & $94.23$ & $94.47$ & $93.06$ & $93.71$ & $94.37$ & $94.55$ \\ 
 $5\%$    & $88.78$ & $90.48$ & $91.97$ & $93.18$ & $93.18$ & $93.18$\\ 
\hline
\end{tabular}
\end{center}
\caption{MNIST Binary results in the form of macro F1 over all classes with various methods and limited training data. }
\label{table:mnist_binary}
\end{table*}  

\subsection{Hierarchical Image Classification}

\textbf{Task} The task here is to classify images of the size $32*32*3$ into $100$ child classes and $20$ parent classes to form a hierarchical structure. This dataset consists of $50$k train examples and $10$k test examples. We randomly select 10\% of the data for the \textbf{low data regime}.

\textbf{Experiments} The main architecture for this task is ResNet50 with Two linear layers on top. 
We have designed two separate models for super-classes and sub-classes. We use the Adam optimizer with a learning rate of $0.001$. We train the models for 20 epochs for the baseline. For primal-dual and sampling loss, we train 20 epochs after ten epochs of training normally. As for the evaluation metric, we use the average of the accuracy values of super-classes and sub-classes. 

The simple Baseline for this task is ResNet18 which provides a smaller model compared to ResNet50 and leads to a drop in a few percentages of accuracy.

\begin{table*}[ht!]
\begin{center}
\begin{tabular}{l ccc c}
\hline
Models            & in-para  & out-of-para & no-effect & Test Acc \\ 
\hline
 Majority      &45.46 &49.47 & 55.0 &30.66 \\
 Adaboost       & 49.41 & 36.61  &48.42  &43.93   \\
 Decomp-Attn      &56.31 &48.56 &73.42 & 59.48\\
\hline
BERT~\cite{Tandon2019WIQAAD}     &79.68 & 56.13 & 89.38 & 73.80 \\
RoBERTa~\cite{Tandon2019WIQAAD}     &74.55 &   61.29 & 89.47 & 74.77 \\
Logic-Guided~\cite{asai2020logic}     & - & - & - & 78.50 \\
RoBERTa + Sampling loss   & 77.19  &  63.28 & 90.25 & 77.65 \\
\textbf{RoBERTa + ILP}     & \textbf{78.55} & \textbf{65.12} & \textbf{90.02} & \textbf{78.98} \\
\hline
  Human     & - & - & - & 96.33\\
\hline
\end{tabular}
\end{center}
\caption{Model Comparisons on WIQA benchmark. The evaluation metric of WIQA test data includes four categories: in-paragraph, out-of-paragraph, no effect, and overall test accuracy. }
\label{table:results_wiqa_main}
\vspace{-3mm}
\end{table*}

\begin{table}[t]
\centering
\begin{tabular}{l}
NLI Consistency Rules \\ \hline
$\top \Rightarrow \operatorname{ent}\left(h_{1}, h_{1}\right)$ \\
$\operatorname{con}\left(h_{1}, h_{2}\right) \Rightarrow \operatorname{con}\left(h_{2}, h_{1}\right)$ \\
$\operatorname{ent}\left(h_{1}, h_{2}\right) \Rightarrow \neg \operatorname{con}\left(h_{2}, h_{1}\right)$ \\
$\operatorname{neu}\left(h_{1}, h_{2}\right) \Rightarrow \neg \operatorname{con}\left(h_{2}, h_{1}\right)$ \\
$\operatorname{ent}\left(h_{1}, h_{2}\right) \wedge \operatorname{ent}\left(h_{2}, h_{3}\right) \Rightarrow \operatorname{ent}\left(h_{1}, h_{3}\right)$
\end{tabular}
\caption{The constraints of the NLI task. $\operatorname{con}(X, Y)$, where $X$ is the hypothesis and $Y$ is the premise is \textit{True} if the $X$ contradicts $Y$. $\operatorname{ent}$ and $\operatorname{neu}$ represent the entailment and undetermined relationship respectively~\cite{minervini2018adversarially}. $ESIM$ \cite{DBLP:journals/corr/abs-1808-08609}}
\label{tab:nli_rules}
\vspace{-5mm}
\end{table}

\subsection{Causal Reasoning}

\textbf{Task} We use WIQA benchmark~\cite{Tandon2019WIQAAD} to evaluate the causal reasoning QA task. In particular, the WIQA benchmark contains $29808$ training samples, $6894$ development samples, and $3993$ test data samples. Formally, the task of the WIQA is to predict an answer $a$ from a set of candidate answers $A \in [$ more, less, no effect $]$ given question $q$ and a document $\mathcal{C}$ that contains sentences $\mathcal{C} = \{c_1, \dots, c_n\}$. Each data sample has a fixed triplet format $(q,C,a)$.
We enforce the constraints of symmetry and transitivity with the models for predicting consistent answers to a set of triplets.
For example, the symmetry constraint is defined as follows:
$symmetric(x) \Rightarrow F(q,C) \land \neg F(\neg q,C)$
where $q$ and $\neg q$ represent the question and its antonym, $C$ represents the same document, and $\neg F$ is the opposite of the answer $F$.

\noindent \textbf{Experiments} 
We use the large-scaled pre-trained language model, RoBERTa, as the backbone architecture that initially followed \cite{asai2020logic}. Besides, we use 2 linear layers of MLP to predict the causal reasoning answer.
We keep $128$ tokens as the max length for the question and $256$ tokens as the max length for the paragraph in each data sample.
We set the batch size of the data to 8 and train the model using 10 epochs.
The learning rate of our model is $1e-2$. 
The model is optimized by Adam optimizer.
Table~\ref{table:results_wiqa_main} shows the model performance on the WIQA benchmark compared to other strong baseline architectures. 
After integrating the constraints of symmetry and transitivity,
we observe that the RoBERTa architecture using sampling loss improves $2.88\%$ over the RoBERTa baseline model. Moreover, the RoBERTa architecture, including ILP as inference, has a $4.21\%$ improvement.



\subsection{Natural Language Inference}
\subsubsection{Task}
Natural Language Inference~(NLI) is the task of evaluating a hypothesis given a premise. The constraints applicable to this task are summarized in Table \ref{tab:nli_rules}.

\subsubsection{Experiments}
For this task, we use a pre-trained RoBERTa base model with two layers of Multilayer perceptron(MLP) to predict a hypothesis given a premise.
To encode the input pair of (Hypothesis, Premise), we concatenate them together and use pre-trained RoBERTa. 
Here, we use an AdamW optimizer with a $1e-5$ learning rate along with cross-entropy loss and set the batch size of the data to 16. The $10\%$ and $100\%$ of training data from SNLI \cite{bowman-etal-2015-large} are used to train the model and train each model for 5 epochs. We use two benchmarks to evaluate the models, SNLI for standard evaluation and $A^{ESIM}_{1000}$ for constraint-focused evaluation. The results from training are shown in table~\ref{table:result_NLI}. According to the table, our experiments show no significant improvement from training using either Primal-Dual (PD) or Sampling loss alone. However, ILP as an inference-time integration significantly improves the accuracy on $A^{ESIM}_{1000}$ around $6-8\%$ on both Primal-Dual and Sampling-loss. We also observe the same effectiveness for using ILP on the model trained with only 10\% of data. Furthermore, Primal-Dual and Sampling-Loss do not help to reduce constrain violations due to the low violation rate on the baseline~($5\%$ violations). Sampling Loss takes around $31.25ms$ for each sample. Meanwhile, Primal-Dual takes up to $43.12ms$ more than baseline. ILP during inference adds around $4ms$ for each sample in each training method. The best method for this task is the combination of Sampling Loss + ILP.

\begin{table*}[ht!]
\begin{center}
\begin{tabular}{ l  c  c  c  c }
\hline
Models & SNIL & SNIL & $A^{ESIM}_{1000}$ & $A^{ESIM}_{1000}$ \\
 & small training & large training & small training & large training \\
\hline
$ESIM$ & - & $87.25$ & - & $60.78$ \\
$ESIM^{AR}$ & - & $87.55$ & - & $73.32$ \\
DomiKnowS without constraints & $88.16$ & $89.65$ & $69.90$ & $71.60$  \\
DomiKnowS with constraints + ILP & $88.65$ & $90.11$ & $77.35$ & $80.25$  \\
Primal-Dual & $88.38$ & $90.04$ & $70.05$ & $73.50$  \\
Primal-Dual + ILP & $88.38$ & $90.04$ & $\textbf{79.90}$ & $79.05$ \\
Sampling Loss & $88.19$ & $90.11$ & $68.65$ & $74.55$ \\
Sampling Loss + ILP & $88.26$ & $\textbf{90.26}$ & $76.30$ & $\textbf{82.20}$ \\
\hline
\end{tabular}
\end{center}
\caption{NLI accuracy results with various methods and limited training data}
\label{table:result_NLI}
\end{table*}


\subsection{Belief Network Consistency}

\textbf{Task} In this task, each example has an entity and the corresponding fact. We concatenate the entity with the facts to form the input text for the models. Each fact can be either $True$ or $False$. 


The Train set, of the size $1.8k$, is small compared to the test set, of size $20k$,  by design. Also, one-third of the Train set is set aside as the dev set. 

\textbf{Experiments} The model we use is RoBERTa-base~\cite{Liu2019RoBERTaAR} topped with two linear layers. In our model, all but the last two transformer layers are frozen. The length of the transformer input is $64$. Since this dataset is easy to solve, we limit the transformer size to make the model less complex. The batch size is 128 to incorporate as many constraints between entities as possible. Here, we use the Adam optimizer with a learning rate of $2e-4$ and train the models for $15$,and $30$ epochs for the data sizes $100\%$,and $25\%$, respectively. The results for various methods are shown in table \ref{table:BB}, in which the primal-dual + ILP method outperforms other methods.

\textbf{Baseline Simple} To create the simple baseline, we use the Word2vec ~(Spacy small) to obtain the representation for each sentence of size 96 and input it to a single layer of MLP. The training parameters are the same as the baseline model except for the learning rate, which is changed to $2e-3$.

\begin{table}[t] 

\centering
\begin{tabular}{ l c c c}
\hline
Model  & $25\%$ & $100\%$ \\ 
 \hline

 Base Domiknows & $94.36$ & $94.90$\\ 
 Base Domiknows + ILP & $93.39$  & $95.11$\\ 
 Sample Loss  & $91.33$  & $94.61$ \\ 
 Sample Loss + ILP & $92.05$  & $96.0$ \\ 
  primal dual & $93.87$ & $95.84$\\ 
 primal dual + ILP & $\textbf{95.43}$ & $\textbf{96.22}$\\ 
\hline
\end{tabular}
\caption{BeliefBank F1-measure results with various methods and limited training data. Since the test data size is large in this case, every small improvement is notable.}
\label{table:BB}
\vspace{-5mm}
\end{table} 

\subsection{Named Entity and Relation Extraction}

\begin{table*}[t]
\centering
\begin{tabular}{cccc|cccc}
\hline
                                           & \multicolumn{3}{c|}{$25\%$ training data}                                                      & \multicolumn{3}{c}{$100\%$ training data}                                                                &                      \\ \cline{2-7}
                                           & \multicolumn{1}{c|}{Entity F1} & \multicolumn{1}{c|}{Relation F1} & Overall F1 & \multicolumn{1}{c|}{Entity F1} & \multicolumn{1}{c|}{Relation F1} & Overall F1           &                      \\ \cline{1-7}
\multicolumn{1}{c|}{Base Model}            & \multicolumn{1}{c|}{$79.86$}     & \multicolumn{1}{c|}{$83.68$}       & $82$         & \multicolumn{1}{c|}{$88.91$}     & \multicolumn{1}{c|}{$91.14$}       & $90.15$                &                      \\
\multicolumn{1}{c|}{Base Model + ILP}      & \multicolumn{1}{c|}{$82.14$}     & \multicolumn{1}{c|}{$\textbf{97.84}$}       & $\textbf{90.86}$      & \multicolumn{1}{c|}{$\textbf{91.77}$}     & \multicolumn{1}{c|}{\textbf{$\textbf{97.84}$}}       & $\textbf{95.14}$                &                      \\
\multicolumn{1}{c|}{Base Model + PD}       & \multicolumn{1}{c|}{$82.12$}     & \multicolumn{1}{c|}{$86.25$}       & $84.42$      & \multicolumn{1}{c|}{$81.3$}      & \multicolumn{1}{c|}{$93.56$}       & $88.12$                &                      \\
\multicolumn{1}{l|}{Base Model + Sampling} & \multicolumn{1}{c|}{$\textbf{83.94}$}     & \multicolumn{1}{c|}{$86.51$}       & $85.36$      & \multicolumn{1}{c|}{$85.90$}          & \multicolumn{1}{c|}{$93$}            & \multicolumn{1}{c}{$89.85$} & \multicolumn{1}{c}{}
\end{tabular}
\caption{The results of applying constraints on the CONLL dataset for the named entity and relationship extraction task.}
\label{tab:conll}
\end{table*}
\begin{table}[ht]
\centering
\begin{tabular}{c c c}
\hline
Relation Type & Argument 1 & Argument 2 \\ \hline
Live In       & Person     & Location   \\ 
Org Base      & Org        & Location   \\ 
Work For      & Person     & Org        \\ 
Kill          & Person     & Person     \\ 
Located In    & Location   & Location  \\
\hline
\end{tabular}
\caption{The domain knowledge about the relationship between relation types and entity types in the CONLL dataset.}
\label{tab:conll_constraints}
\end{table}

\subsubsection{Task}
The named entity and relation extraction task is designed to evaluate the models' performance in detecting the entity types and their relationships in a document.
To simplify the task setting, we ignore the span recognition task and consider phrases as given inputs. We further limit the candidates for relationship classification and only classify given pairs of entities based on the relationship that exists between them.
Respectively, there is no `None' class for the relationship classification task. The types of entities are `person', `organization', `location', and `other'. The possible types of relationships are `live-in', `work-for', `located-in', `kill', and `orgbase-on'.
Table \ref{tab:conll_constraints} summarizes the existing relationship constraints between relation types and entity types. Other constraints are the mutual exclusivity between different named entity types for each mention or formally defined as: 
\begin{align*}    
&\forall x \in entities \\ & \text{IF } \operatorname{Type_{A}}(x) \Rightarrow \neg \lor_{Type_{B} \in \text{entity\_types}}^{Type_{B} \neq \operatorname{Type1}} \operatorname{Type_{B}}(x),
\end{align*}
where $\operatorname{Type_{A}} \in \text{entity\_types}$ and  $\operatorname{Type_{Y}}(x)$ is $True$ if $x$ is of type $Y$. A similar mutual exclusivity constraint is also defined over the relationship types for each pair of entities.
\subsubsection{Experiments}
In the base model, we represent each token in the document using a pre-trained BERT-base model~\cite{devlin2018bert}and a pre-trained Word2Vec model~\cite{spacy2}. We concatenate these two representations and take the mean over the tokens of each span to represent it. Then we feed these vectors to boolean classifiers for each entity class. To classify the relationships, we concatenate the representations of two entities and feed them to the boolean classifiers for each relationship type. As the CONLL2003 dataset does not provide train/validation/test splits, we randomly create those sets and further subsample the training set with 20\% of the data to generate a low-resource setting.  To create a simpler model as a baseline, we remove the BERT-base representation and add a Bi-LSTM layer after getting the representations from the Word2Vec from Spacy.  For all experiments, the learning rate is set to $1e-2$, and they are reported on the best performing model on the dev set~(in terms of macro-F1) as we trained the models for 200 epochs. All models are optimized by Adam optimizer.

\noindent\textbf{Baselines:} Pre-trained Bert + Spacy W2V + 1 Layer MLP 
\textbf{Baseline Simple:} Spacy W2V + LSTM + 1 Layer MLP

The detailed results of our baselines performance on both entity and relationship detection tasks are listed in Table \ref{tab:conll}. The results indicate the constraint integration methods are majorly biased toward a better relationship extraction. This is perhaps due to the fact that the relationship pairs have been limited to the set of known pairs, and the base performance of the relationship classifier is higher than the entity recognizer.

\subsection{BIO Tagging}
\subsubsection{Task} 
we select the CONLL-2003 Shared Task benchmark to evaluate the BIO tagging task. The CONLL benchmark contains $14987$ training samples, $3466$ development samples, and $3684$ test samples, including $9$ tagging labels. 
The evaluation metrics in this task include Precision, Recall, and F1.
Formally, the process of computing the score for a partial tagging sequence in the training time is calculated as follows: 
\begin{align*}
    f(y_{[1,\dots,T]}, W) = \sum_{i=1}^Tlogp(y_i|W),    
\end{align*}
where $T$ is the sequence length, $W$ is the learnable weight.
We use $A*$ search over tag prefixes during the inference time to integrate constraints in the BIO tagging outputs. To integrate BIO constraints, Specifically, we add the constraint function $c \in C$, where 
$C$=$\{$`B-*' $\prec$ `I-*', `O' $\not \prec$ `I-*' $\}$
, $\prec$ represents the `before' relation, and $\not \prec$ represents the `not before' relation.
Formally, the process of a constrained tagging sequence in the inference time is computed as follows: 
\begin{align*}
f(y_{[1,\dots,T]}, &W) =\\& \sum_{i=1}^Tlogp(y_i|W)-\sum_{c \in C} c(y_{[1,\dots,T]}, W),
\end{align*}
where $T$ is the sequence length and $W$ is the learnable weight.

\subsubsection{Experiments} We separately apply BI-LSTM and BERT as backbone following a $2$-layers of MLP to predict the sequence tagging for each sentence. We select the cross entropy as the loss function to train the model. The model is optimized by Adam optimizer while the learning rate of our model is $1e-3$.
We set the batch size of the data to $64$ and train the model using $20$ epochs.
Moreover, in the inference time, we use $A*$ search and ILP over tag prefixes to integrate constraints on the BIO tagging outputs. We use both small data scales ($30\%$,  $60\%$) and full data ($100\%$) to train and evaluate the model performance. The results are shown in Table~\ref{table:results_bio_tag}.
The model takes $20$ epochs after initializing the parameters with the BIO model. In the inference time, the model integrating $A*$ decoding has a $2.25\%$ improvement on small data and $0.74\%$ improvement on full data compared to the baseline. Furthermore, including ILP as inference on the top of the model improves $0.72\%$ over the baseline model on full data.
At the same time, we evaluate the time complexity between different models. The baseline model takes $1.01 ms$ for each sample. The model integrating $A*$ constraints takes $1.32 ms$, while the model integrating ILP constraints takes $1.34 ms$. It shows that the inference time complexity has not increased significantly.

\begin{table}[ht!]
\small
\begin{center}
\begin{tabular}{c l ccc}
\hline
Data & Models  & Precision  & Recall & F1   \\ 
\hline
$30\%$ & Bi-LSTM     & 78.86 & 76.34 & 77.58  \\
      & Bi-LSTM+BIO+A$*$     & 81.75 &  78.01 & 79.84  \\
      & Bi-LSTM+BIO+ILP     & 81.68 & 78.21 & 79.65 \\
      & BERT & 84.12  & 78.43 & 81.17 \\
      & BERT+BIO+A$*$ & 81.87 & 86.50 & 84.12 \\
      & BERT+BIO+ILP & 81.76 & 86.68 & 84.15 \\
\hline
$60\%$ & Bi-LSTM     & $78.03$ & $80.46$ & $79.93$  \\
      & Bi-LSTM+BIO+A$^*$     & $79.92$ &  $83.55$ & $81.70$  \\
      & Bi-LSTM+BIO+ILP     & $79.83$ & $83.55$ & $81.65$ \\
      & BERT & $87.80$ & $87.76$ & $87.78$ \\
      & BERT+BIO+A$^*$ & $93.21$ & $84.08$ & $88.40$ \\
      & BERT+BIO+ILP & $93.24$ & $84.06$ & $88.41$ \\
\hline
$100\%$ & Bi-LSTM     & $86.25$ & $84.68$ & $85.46$ \\
      & Bi-LSTM+BIO+A$^*$     & $87.26$ &  $85.17$ & $86.20$  \\
      & Bi-LSTM+BIO+ILP     & $87.28$ & $85.14$ & $86.18$ \\
      & BERT & $93.02$  & $88.87$ & $90.90$ \\
	  & BERT+BIO+A$^*$ & $\textbf{93.64}$ & $89.77$ & $91.66$ \\
      & BERT+BIO+ILP & $93.59$ & $\textbf{89.87}$ & $\textbf{91.70}$ \\
\hline
\end{tabular}
\end{center}
\caption{Model Comparisons on CONLL-2003 BIO-tagging test benchmark. A$^*$ is A$^*$ decoding.}   
\label{table:results_bio_tag}
\end{table}



\subsection{Arithmetic Operation with Supervision}
\subsubsection{Task}
The goal of the MNIST Arithmetic task is to train an MNIST digit classifier with the only supervision being the sum of digit pairs. Constraints for such a task would be to produce predictions whose sum matches the given sum:
\begin{align*}
    &S(\{img_1, img_2\}) \Rightarrow \\
    &\bigvee_{M=max(0,S-9)}^{M=min(S,9)} M(img_1) \wedge \{S-M\}(img_2),
\end{align*}
Where $S(\{img_1, img_2\})$ indicates that the given summation label is $S$ and $M(img_i)$ indicates that the $i$th image has the label $M$.

\subsubsection{Experiments}
\label{appendix:mnistaarithresults}
Constraint violation or satisfaction corresponds to the rate at which the constraints for each sample are violated or satisfied, averaged across all samples. The constraint satisfaction rate and accuracy rate are comparable between each setting, with a slight drop in percentage points due to the fact that an image pair may predict correctly 1 out of 2 digits but still not fully satisfy the constraints. \\
The full results for accuracy and constraint violation are in Table \ref{table:digitArithmeticResults} and Table \ref{table:digitArithmeticConstraints}.

\begin{table}[ht!]
\begin{center}
\begin{tabular}{ l c c }
\hline
Setting & Small Data & Large Data \\ 
\hline
Digit Labels & $94.62$ & $98.54$ \\
Primal-Dual & $95.33$ & $98.40$ \\
Sampling Loss & $\textbf{95.92}$ & $98.56$ \\
Semantic Loss & $95.12$ & $\textbf{98.62}$ \\
Explicit Sum & $94.93$ & $98.55$ \\
Explicit Sum + ILP & $94.93$ & $98.55$ \\
Baseline & $10.32$ & $9.01$ \\
Baseline + ILP & $10.32$ & $6.90$ \\
\hline
\end{tabular}
\end{center}
\caption{Accuracy on various constraints/data settings for the MNIST Arithmetic task.}
\label{table:digitArithmeticResults}
\end{table}

\begin{table}[t]
\begin{center}
\begin{tabular}{ l c c }
\hline
Setting & Small Data & Large Data \\ 
\hline
Digit Labels & $10.24$ & $2.86$ \\
Primal-Dual & $12.52$ & $3.18$ \\
Sampling Loss & $7.96$ & $2.86$ \\
Semantic Loss & $9.44$ & $2.76$ \\
Explicit Sum & $9.78$ & $2.88$ \\
Baseline & $\textbf{94.50}$ & $\textbf{96.92}$ \\
\hline
\end{tabular}
\end{center}
\caption{Constraint violation rate for the MNIST Arithmetic task in the small and large data setting.}
\label{table:digitArithmeticConstraints}
\end{table}

For the digit classifiers, we use a simple LeNet-style CNN architecture \cite{lecun1998gradient}. 
The "explicit sum" model, in order to compute the summation probability distribution $P(S = s)$, sums the Softmax distributions of the two digits $P(D_1)$ and $P(D_2)$. i.e.
\begin{align*}
  &P(S = s) = \\ &\sum^{min(s,9)}_{k = max(0,s-9)} P(D_1 = k) P(D_2 = s - k)  
\end{align*}
We compare Primal-Dual and Sampling Loss with three other methods: 1) A baseline model consisting of a two-layer MLP on top of the digit classifier logits, supervised on the summation values instead of using constraints. 2) Training with digit labels directly (i.e. regular digit classification). 3) Explicitly specifying the summation constraints through the model architecture by directly computing the predicted summation value from the digit classifier logits (the "explicit sum" model). On both the small and large data settings, training with Primal-Dual, Sampling Loss, or Semantic Loss using constraints as the only source of supervision performs almost identically to both direct supervision on digit labels and the explicit sum model. On the other hand, training with neither direct supervision nor constraints performs around random guessing in both the small and large training data cases. ILP does not improve the "explicit sum" case as it already achieves a high rate of constraint satisfaction, and it hurts the baseline case slightly as the baseline model probabilities are not informative. Similar results can be seen when looking at constraint satisfaction scores in the large data setting. In the small data setting, there's slightly more variability with Primal-Dual reaching a 12.52\% constraint violation rate and Sampling Loss having just a 7.96\% rate of constraint violation.

\subsection{Sudoku}
\subsubsection{Task}
The task of Sudoku is to predict the missing numbers from a $n*n$ Sudoku table such that the final table is valid according to the following rule:
"There should not be two cells from each row/block/column, with the same value" or formally defined as:
\begin{align*}
    \text{IF } &\operatorname{digit}(x, i) \land \\ &(\operatorname{same\_row}(x, y) \lor \\&\operatorname{same\_col}(x,y) \lor \\ &\operatorname{same\_block}(x,y)) \\
    & \Rightarrow \neg \operatorname{digit}(y,i),
\end{align*}
where $x$ and $y$ are variables regarding the cells of the table, $i \in [0, n]$ for a $n*n$ Sudoku, $digit(x,i)$ is $True$ only if the value of $x$ is predicted to be $i$.

For this task, we use an incomplete $9*9$ Sudoku as the harder task~(large-data) and $6*6$ Sudoku representing a simpler task~(low-data).

\subsubsection{Experiments}
Here, we use a simple learnable vector of size $n*n*n$, which stores the probability of each $n$ assignments given all the cells of the table, i.e. if $0.6$ is present in (1, 1, 2) index of the vector, the probability of row 1, column 1 being number $2$ is predicted to be 60\%.
We train this vector using two available sources of supervision: 1) The input of the Sudoku through a CrossEntropy loss function. 2) The indirect loss function based on constraints when the input is masked.
All models have been trained with a learning rate of $1e-1$ and use the SGD optimizer. Since the objective is to complete the Sudoku with correct numbers, we continue the training until the goal is accomplished or the $250$ epochs have passed.

\end{document}